\documentclass{article}

\usepackage[dblblindworkshop,final]{neurips_2026}

\usepackage[utf8]{inputenc}
\usepackage[T1]{fontenc}
\usepackage{hyperref}
\usepackage{url}
\usepackage{booktabs}
\usepackage{amssymb}
\usepackage{amsmath}
\usepackage{microtype}
\usepackage{xcolor}
\usepackage{graphicx}
\usepackage{multirow}
\usepackage{enumitem}
\usepackage{listings}

\usepackage{fontawesome5}
\usepackage{hyperref}

\definecolor{codekw}{HTML}{356A96}
\definecolor{codecomment}{HTML}{5E6872}
\definecolor{codestr}{HTML}{D88032}
\definecolor{codebad}{HTML}{B64B4B}
\definecolor{codegood}{HTML}{397A5A}
\lstdefinestyle{pysnippet}{
  language=Python,
  basicstyle=\scriptsize\ttfamily,
  keywordstyle=\color{codekw}\bfseries,
  commentstyle=\color{codecomment}\itshape,
  stringstyle=\color{codestr},
  showstringspaces=false,
  columns=fullflexible,
  keepspaces=true,
  aboveskip=2pt,
  belowskip=0pt,
  escapechar=@,
}

\newcommand{\gv}{Agent-as-a-Verifier}
\newcommand{\lav}{LLM-as-a-Verifier}
\newcommand{\statusv}[1]{\textsc{#1}}

\title{Agent-as-a-Verifier: From Scores to Evidence for\\ General Verification}

\workshoptitle{Who Verifies the Agents? Toward Reliable Agent Development}

\author{%
  Yuxuan Sun$^{1}$\quad
  Hao (Mark) Chen$^{1}$\quad
  Nagarajan Natarajan$^{2}$\quad
  Sushrut Karmalkar$^{2}$\\
  \textbf{Niranjani Prasad}$^{2}$\quad
  \textbf{Amit Sharma}$^{2}$\quad
  \textbf{Hongxiang Fan}$^{1}$\\[0.6em]
  $^{1}$Imperial College London\quad
  $^{2}$Microsoft Research
}

\begin{document}

\maketitle

\begin{abstract}
Verification is increasingly critical across various inference and learning paradigms, such as test-time scaling, recursive intelligence and automated research, where feedback can improve efficiency and optimization.
Common score-based verifiers, or discriminative reward models, predict a scalar reward for an output.
While useful for filtering, rewarding, and ranking, such scores provide no explicit, auditable evidentiary basis for the judgment. 
In parallel, although existing generative process reward models offer text feedback, most designs do not fully exploit agentic capabilities, such as tool use and iterative reasoning, which are essential for fact-checking and fine-grained claim verification.
In this paper, we address these limitations by introducing \emph{\gv{}}, an agentic, certificate-producing
verification system that places role-separated LLM agents inside a
formal-methods-inspired claim--obligation--certificate protocol.
Given a generation candidate, \gv{} produces a certificate by extracting a verdict-blind dependency graph of claims,
independently compiling the task's acceptance contract, and deriving checkable obligations.
On Terminal-Bench~2.1, \emph{\gv{}} achieves higher accuracy than the state-of-the-art approach. 
Across different downstream applications, the augmented agent also outperforms the plain agent on FrontierMath and CyberGym. More importantly, \emph{\gv{}} is able to identify reward-hacking solutions that deceive the benchmark evaluator, highlighting its broader potential for iterative and recursive agent optimization. 

\vspace{0.3em}
\noindent\faGithub\ \textbf{Code:} \href{https://github.com/sunyxedu/General-Verifier}{\texttt{github.com/sunyxedu/General-Verifier}}

\end{abstract}

\section{Introduction}
\label{sec:intro}

Verification has become a central capability throughout the language-model lifecycle~\citep{kwok2026lav}. 
Across diverse learning and inference paradigms, it can provide reward signals during training, guide search and decision-making during inference, and support evaluation and auditing after deployment. Unlike prior work that often treats verification as a mechanism tied to specific settings such as test-time scaling or completed rollouts, we view this as a general verification  problem.
Toward this goal, we focus on evidence-backed verification with dedicated agents: a general
verifier can verify completed traces and outcomes or steer an iterative agent-based execution or optimization.

As a first step toward our long-term vision, this work focuses on test-time scaling (TTS), where additional inference-time computation is used to improve model performance, for example through sampling, search, or iterative refinement~\citep{snell2024scaling, chen2026rethinking}.
A common approach of TTS is best-of-$N$ selection, in which multiple candidate solutions are sampled and a verifier assigns scalar rewards to select the most promising one.
Sampling more candidate solutions increases the chance that at least
one is correct, but realizing that gain depends on selecting the right
candidate~\citep{brown2024large,snell2025scaling, zhao2025sample}. 
Recent methods
therefore use LLMs to score and rank candidate runs: \lav{} \citep{kwok2026lav} scores candidates
against task criteria and can report progress during execution. BoN-MAV combines votes on several aspects
\citep{lifshitz2025mav}. Weaver combines several noisy verifier signals
\citep{saad2026weaver}.
However, these trajectory-scoring interfaces infer correctness from the supplied
representation rather than independently re-checking the delivered state or
preserving replayable evidence for the verdict. They also inherit known
sensitivities to presentation, ordering, adversarial rewriting, and model
self-preference \citep{wang2023fair, zheng2023judging, gu2024survey,
wu2023style, zheng2025cheating, khalifa2026gaming, panickssery2024llm,
goel2025great}.

While existing tool-grounded verifiers do execute checks, they are typically
designed for revision \citep{gou2024critic}, domain-specific formal checking
\citep{zhou2024dtv, liu2025safe}, or graphs supplied in advance
\citep{fang2026gov}, rather than post-hoc selection among unmodified agent
traces. We therefore ask whether a verifier can reconstruct checkable claims
from such traces, inspect the delivered environment, preserve the resulting
evidence, and abstain when a claim cannot be established.

We present the \textbf{\gv{}}, a hybrid verification system that combines the
adaptability of role-separated LLM agents with formal-methods-inspired
structure. LLM agents perform semantic decomposition and propose, execute, and
audit decision procedures.
Central to this framework are claim DAGs, explicit obligations, typed provenance,
replay artifacts, evidence-strength rules, and fail-closed folds,
which together constrain how
those procedures become verdicts. 
Individual checks may be formal, symbolic,
or executable, but the system is not an end-to-end proof system: semantic
compilation and evidence auditing remain model-dependent. Its organizing
principle is to separate LLM proposals about \emph{what to check} from
replayable evidence about \emph{what the checks found}.
Figure~\ref{fig:arch} summarizes the evaluated selection workflow in three
stages. In summary, our contributions include:

\begin{itemize}[leftmargin=2em]
  \item We frame agent verification as certificate construction rather than
    scalar scoring, and define a hybrid agentic-symbolic interface in which
    role-separated LLM agents propose, execute, and audit formal, symbolic, and
    other executable checks under explicit claims, obligations, evidence
    provenance, and abstention.
  \item We introduce \gv{}, a general verification framework that imports ordinary
    rollouts into a digest-bound event ledger, extracts a verdict-blind claim
    DAG, compiles the final task contract,
    and separates check proposal, execution, and evidence audit. A lexicographic rule ranks certificates by verdict and verified contract coverage, without task-specific verifier training.
  \item We evaluate \gv{} on Terminal-Bench~2.1, where it achieves higher accuracy than the state-of-the-art approach in best-of-$N$ selection, and in the loop of a working agent on FrontierMath, Terminal-Bench~2.1, and CyberGym, where it improves the same agent on all three. More importantly, \gv{} identifies reward-hacking solutions that deceive the benchmark evaluator, highlighting its potential for iterative and recursive agent optimization.
\end{itemize}

\section{Background and Related Work}
\label{sec:judges}

\subsection{Background}

Test-time scaling improves model performance by allocating additional computation at inference time. Broadly, this can take a {sequential} paradigm~\citep{guo2025deepseekr1, muennighoff2025s1}, in which a model iteratively reasons, revises, or interacts with an environment, or a {parallel} paradigm~\citep{brown2024large, snell2024scaling, chen2026fasttts}, in which multiple candidate solutions are generated and a verifier or selection rule chooses among them. Both forms become increasingly important in inference-time optimization, where computation can be spent not only on generating longer reasoning traces but also on exploring alternative actions and outcomes.

In recent agentic task, an agent run produces a sequence of tokens and tool-call returns, together with
a delivered result, such as a final answer or a modified workspace.
A score-based
verifier maps an answer or recorded run to a scalar that can supervise training
or rank candidates in best-of-$N$ selection. Outcome and process reward models
established this pattern for sampled solutions
\citep{cobbe2021training,lightman2024let,uesato2022solving}. 
Here, we use
\emph{trajectory} only for the recorded tokens and tool-call returns supplied to a
trajectory-scoring method, rather than the delivered result.

\subsection{Related Work}

\paragraph{Score-based verification.}
Recent training-free methods rank candidate runs through pairwise judgments,
multiple verifier votes, or combinations of noisy verifier signals
\citep{kwok2026lav,lifshitz2025mav,saad2026weaver}. These methods differ
in how they produce a score, and \lav{}~\citep{kwok2026lav} can also report progress during a run,
but their usual selection interface does not retain a checkable account of what
was verified.

\paragraph{Evidence-producing verification.}
Other work supplies parts of a richer interface. FActScore and FacTool split an
output into smaller claims and check them against retrieved or tool-produced
evidence \citep{min2023factscore,chern2023factool}. Premise- and graph-based
methods verify links between reasoning steps
\citep{ling2023deductive,fang2026gov}, while formal systems
translate informal reasoning into domain-specific proofs or symbolic checks
\citep{zhou2024dtv,liu2025safe,saldivar2026theoria}. Closer to tool-using
agents, Agent-as-a-Judge and Gandalf inspect workspaces, artifacts, or tool state
\citep{zhuge2024agent,ducker2026verifying}; other systems use tools or deliberate
over checkable subclaims \citep{sedoc2026trust,xu2026tirjudge,zhang2026agentvrl}.
These capabilities are not individually new. We study their combination in one
verifier for ordinary agent runs: extract claims, derive final requirements
independently, run checks, review and save the evidence, and report ``not
verified'' when a reliable check is unavailable.

\subsection{Motivation and Analysis}
\begin{figure}[t]
  \centering
  \includegraphics[width=\textwidth]{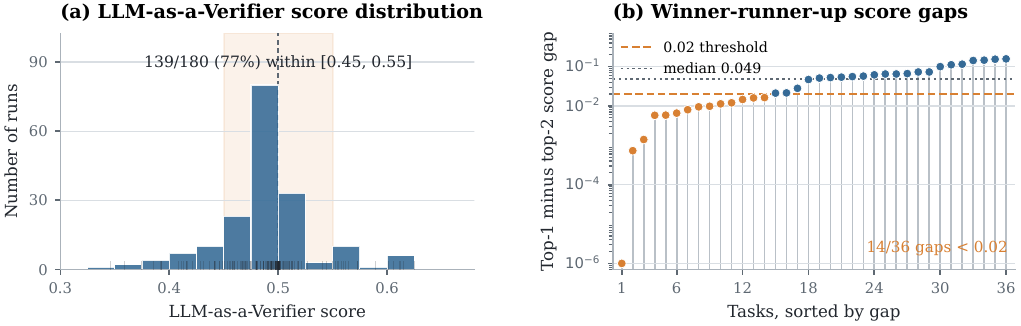}
  \caption{\lav{} scores (\texttt{gpt-5.6-luna}) on 180
  candidate runs from the 36 mixed-outcome Terminal-Bench~2.1 task pools of
  \lav{}'s own released dataset (five runs per task).
  \textbf{(a)} Candidate-score distribution; 139/180 scores fall within
  $[0.45,0.55]$, and all 180 lie in $[0.346,0.625]$. \textbf{(b)} For each task,
  the highest score minus the second-highest score, sorted for display on a
  log scale. The median gap is $0.049$, and 14/36 gaps are below $0.02$.}
  \label{fig:lav-score-gaps}
\end{figure}

Figure~\ref{fig:lav-score-gaps} shows that many \lav{} scores and top-two margins
are tightly clustered. These scores rank candidates but do not say what was
checked or retain the evidence behind the choice. 
Separately, a recorded trajectory may omit final-state evidence required for tool-using tasks, forcing the verifier to judge only what the agent recorded. Such judgments can therefore be sensitive to presentation changes, hidden bugs, and model-family preferences
\citep{khalifa2026gaming,zhou2026convincing,liao2026redteaming,
panickssery2024llm,goel2025great}.
Direct checks do not guarantee correctness,
but tests, file inspection, and proof tools provide evidence beyond the agent's
wording \citep{stechly2025self}. This motivates four goals: state the
requirements, check them, save the evidence, and report failed or unavailable
checks as ``not verified.''

\begin{figure}[t]
  \centering
  \includegraphics[width=\textwidth]{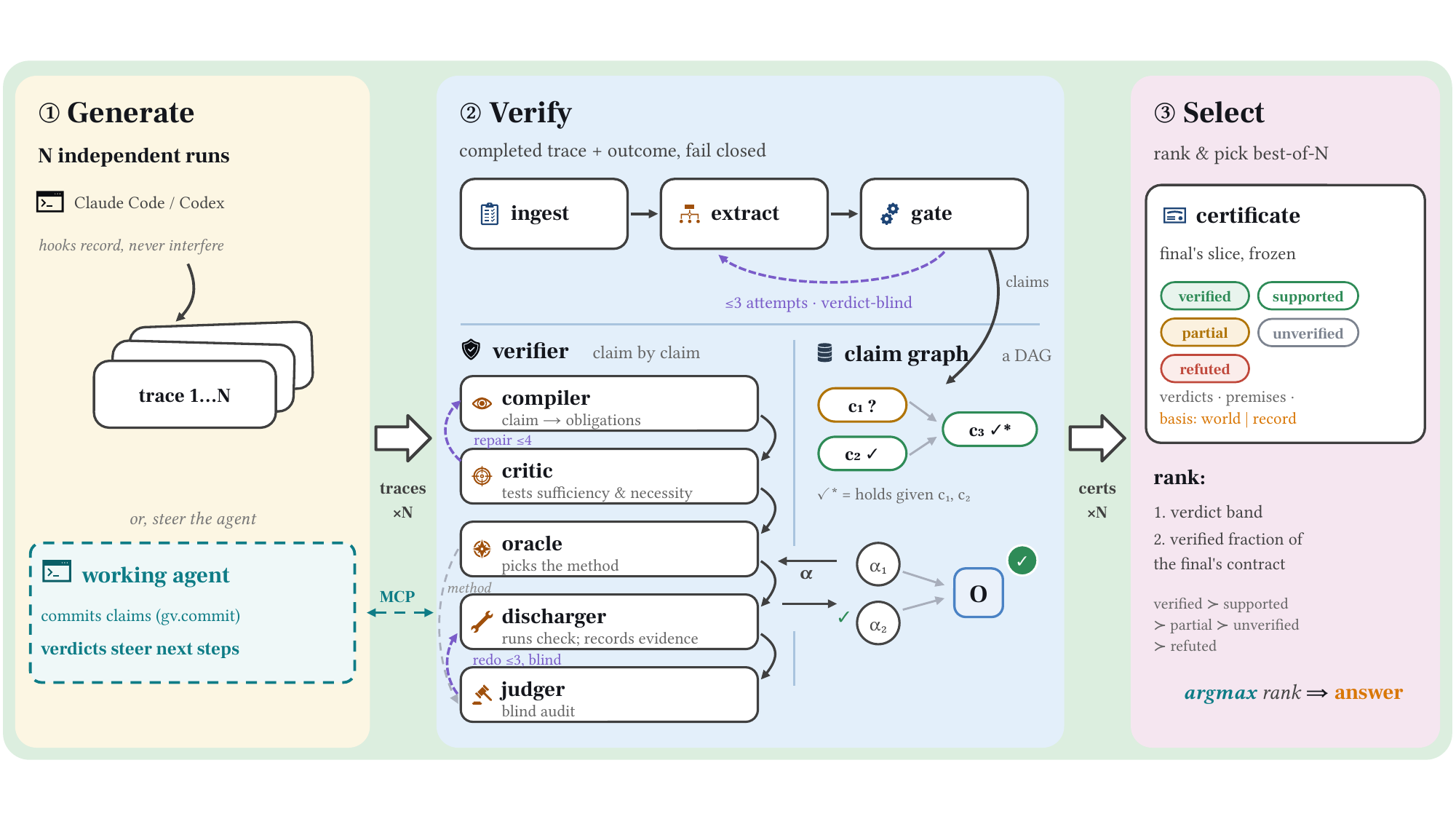}
  \caption{The evaluated \gv{} workflow has three stages:
  (1) generate and record $N$ candidate runs; (2) turn each run into claims,
  run and review the required checks, and save the result as a certificate;
  and (3) select the strongest certificate. The dashed path shows the same
  verifier returning failed checks or counterexamples while an agent is still
  working (App.~\ref{app:steering}).}
  \label{fig:arch}
\end{figure}

\section{Agent-as-a-Verifier}
\label{sec:method}

\paragraph{Framework overview.}
Figure~\ref{fig:arch} presents an overview of generate--verify--select workflow.
In the generation phase, each candidate's record and delivered
environment are collected before verification starts. Verification is then
organized around a \emph{separation of authority}: a deterministic
orchestrator coordinates six stateless LLM roles, alone owns the verification
state, enforces the protocol, and computes every verdict. LLM roles may
propose what should be claimed, checked, or executed, or audit the resulting
evidence, but no LLM role can directly award a passing verdict. This division
is intended to preserve strictness even when an individual role is incomplete
or mistaken.

The roles are arranged in three stages, described in turn below:
(i) an \emph{extractor} turns the candidate into a graph of claims, an
\emph{obligation compiler} turns each claim into required checks, and a
\emph{decomposition critic} challenges those checks (\S\ref{sec:extract});
(ii) for each check, an \emph{oracle} proposes a procedure, an
\emph{executor} runs it, and a \emph{judger} audits the result
(\S\ref{sec:game}); and (iii) the orchestrator folds the audited evidence
into a verdict and a rank (\S\ref{sec:select}). Information is deliberately
asymmetric across roles to limit hindsight and self-confirmation: for example,
the extractor never sees check results, and the compiler of the final task
contract never sees the candidate's own account. Roles exchange only
schema-validated objects, not shared conversational state; crashes, malformed
outputs, unavailable checks, and failed audits contribute no positive
evidence. Detailed contracts, sandboxing, and retry rules appear in
App.~\ref{app:roles}.

\subsection{Inputs and Certificate}
\label{sec:task-description}

For a completed candidate, let
\[
X=(q,R,y,W,W_0)
\]
denote the task ($q$), execution record ($R$), final answer ($y$), delivered
workspace ($W$), and optional pre-run snapshot ($W_0$), as shown
in~Fig.~\ref{fig:arch}. These inputs identify a fixed candidate, but
verification is active: checks may inspect files and execute programs in the
captured environment. The verifier returns the certificate
\begin{equation}
\Gamma=\mathcal V(X)=\bigl(G,\mathcal O,\mathcal E,v_\star\bigr),
\label{eq:certificate}
\end{equation}
where $G$ is a claim graph (\S\ref{sec:extract}), $\mathcal O$ the required
checks for its claims (\S\ref{sec:extract}), $\mathcal E$ the reviewed
evidence for those checks (\S\ref{sec:game}), and $v_\star$ the verdict on the
delivered outcome (\S\ref{sec:select}). The certificate doubles as an
auditable report.

\subsection{Claims and Required Checks}
\label{sec:extract}
\label{sec:critic}

\paragraph{Claim graph.}
The candidate is rewritten as a DAG
\begin{equation}
G=(C,D,c_\star)=\operatorname{Extract}(q,R,y),\qquad
C=C_{\mathrm{A}}\cup C_{\mathrm{E}}\cup\{c_\star\},
\label{eq:extract}
\end{equation}
where $(c' \rightarrow c)\in D$ means that the argument for $c$ relies on $c'$.
\emph{Effects} $C_{\mathrm E}$ are generated mechanically from $R$: each is a
reference $(\text{event},\text{tool},\text{digest})$ to one recorded tool call
and its result, so a claim about what happened is bound to the record rather
than to the agent's summary. The extractor (read-only) adds \emph{assertions}
$C_{\mathrm A}$, atomic falsifiable statements that may cite effects as
premises, and the final claim $c_\star$, the delivered outcome. It runs before
any check, so evidence cannot change what is evaluated.

\paragraph{Obligations.}
An obligation is a triple $o=(\varphi_o,\beta_o,\sigma_o)$: a checkable condition $\varphi_o$, the minimum evidence basis
$\beta_o\in\{\textsf{record}\prec\textsf{world}\}$ that may settle it (\textsf{world}: established against the delivered state; \textsf{record}: shown only by the recorded tool calls in $R$), and a severity
$\sigma_o\in\{\textsf{must},\textsf{should}\}$ stating whether the task makes
acceptance depend on $\varphi_o$. Each claim receives a set intended to
satisfy $\bigwedge_{o\in\mathcal O(c)}\varphi_o\Rightarrow c$:
\begin{equation}
\mathcal O(c)=
\begin{cases}
\operatorname{Plan}(q,c,G;R,W), & c\in C_{\mathrm A},\\
\operatorname{Match}(c,R), &c\in C_{\mathrm E},\\
\operatorname{Contract}(q,W_0), & c=c_\star.
\end{cases}
\label{eq:obligations}
\end{equation}
$\operatorname{Plan(.)}$ checks one step of the candidate's argument and may consult the graph, record, and workspace. 
Dependencies are handled by the fold
(Eq.~\eqref{eq:premise}). 
$\operatorname{Match(.)}$ yields a single \textsf{record} obligation: the digest referenced by $c$ equals the digest recomputed from the corresponding tool call in $R$.
$\operatorname{Contract(.)}$ compiles the task's acceptance criteria with no
tools from $q$ and pre-task baseline metadata alone: it sees neither $G$, $R$,
$y$, nor $W$, so a candidate cannot narrow its own goal (e.g., to ``the
provided tests pass'').

\paragraph{Decomposition critic.}
For $c\in C_{\mathrm A}\cup\{c_\star\}$, a non-executing critic attacks
$\mathcal O(c)$ in both directions by proposing a state $s$ with
\begin{equation}
\underbrace{s\models\textstyle\bigwedge_{o}\varphi_o\ \wedge\ s\not\models c}_{\text{insufficient}}
\qquad\text{or}\qquad
\underbrace{s\models c\ \wedge\ \exists o:\ s\not\models\varphi_o}_{\text{over-strict}}.
\label{eq:critic}
\end{equation}
The first reveals a missing condition, while the second reveals a condition that would refute a correct candidate. 
Over critic rounds $k=1,\dots,K$, a standing
attack after round $k<K$ triggers a repair
$\mathcal O^{(k+1)}(c)=\operatorname{Repair}(\mathcal O^{(k)}(c),s)$, and the
critic, which sees the full history, is instructed not to reissue an attack
it has overruled. If an attack still stands after round $K$, $c$ stays
\statusv{unverified}. These are model-driven challenges.

\subsection{Checking and Reviewing Evidence}
\label{sec:game}
\label{sec:basis}

Each surviving obligation is proposed (oracle), run (executor), and reviewed
(judger) by separate roles:
\begin{equation}
\pi_o=\operatorname{Propose}(o),\qquad
r_o=\operatorname{Run}(\pi_o;R,W,W_0),\qquad
e_o=\operatorname{Review}(o,\pi_o,r_o)=(z_o,b_o,\ell_o,a_o).
\label{eq:check}
\end{equation}
$\pi_o$ may use tests, file inspection, symbolic calculation, external lookup,
or a proof tool; $\operatorname{Review}$ audits sufficiency and execution
fidelity and may order bounded re-runs. In $e_o$,
$z_o\in\{\emph{holds},\emph{refuted},\emph{unavailable},\emph{failed}\}$ is
the outcome (\emph{unavailable}: not checkable in the captured environment;
\emph{failed}: the check or review did not complete),
$b_o\in\{\textsf{record},\textsf{world}\}$ the basis achieved,
$\ell_o\in\{\textsf{limited}\prec\textsf{conclusive}\}$ the strength
(App.~\ref{app:ladder}), and $a_o$ a replayable witness. Evidence counts only
if it is admissible,
\begin{equation}
b_o\succeq\beta_o
\quad\text{and}\quad
\bigl(z_o=\emph{refuted}\Rightarrow\ell_o=\textsf{conclusive}\bigr);
\label{eq:admissible}
\end{equation}
otherwise, and on any crash, timeout, malformed output, or failed review,
$z_o$ is set to \emph{failed}. In particular, record evidence can neither
satisfy nor refute a \textsf{world} obligation. 

\subsection{Certificate Verdict and Selection}
\label{sec:select}

The orchestrator, not an LLM role, folds evidence into verdicts. A violated
\textsf{should} clause is a contrary signal, not a counterexample, so let
$\tilde z_o=\emph{unavailable}$ if $z_o=\emph{refuted}\wedge\sigma_o=\textsf{should}$,
and $\tilde z_o=z_o$ otherwise. The local verdict of a claim is
\begin{equation}
v_{\mathrm{loc}}(c)=
\begin{cases}
\statusv{refuted}, & \exists o:\ \tilde z_o=\emph{refuted},\\
\statusv{verified}, & \mathcal O(c)\neq\emptyset\ \wedge\ \forall o:\ \tilde z_o=\emph{holds},\\
\statusv{partial}, & \forall o:\ \tilde z_o\in\{\emph{holds},\emph{unavailable}\}\ \wedge\ \exists o:\ \tilde z_o=\emph{unavailable},\\
\statusv{unverified}, & \text{otherwise (e.g., any \emph{failed} check)}.
\end{cases}
\label{eq:fold}
\end{equation}
With $\statusv{refuted}\prec\statusv{unverified}\prec\statusv{partial}\prec\statusv{verified}$,
an intermediate claim takes the weakest status along its dependencies:
\begin{equation}
v(c)=\min\Bigl(v_{\mathrm{loc}}(c),\ \min_{(c',c)\in D}v(c')\Bigr),\qquad c\ne c_\star .
\label{eq:premise}
\end{equation}
The final claim is exempt: $\mathcal O(c_\star)$ is the independently compiled
task contract, so a broken step in the candidate's argument removes a line of
support but does not by itself refute the delivered outcome. Its verdict
refines $v_{\mathrm{loc}}(c_\star)$ with two rules,
\begin{equation}
v_\star=
\begin{cases}
\statusv{unverified}, & v_{\mathrm{loc}}(c_\star)=\statusv{verified}\ \wedge\ \forall o:\ b_o=\textsf{record},\\
\statusv{supported}, & v_{\mathrm{loc}}(c_\star)=\statusv{verified}\ \wedge\ \exists o:\ b_o=\textsf{world}\ \wedge\ \exists o:\ \ell_o=\textsf{limited},\\
v_{\mathrm{loc}}(c_\star), & \text{otherwise}
\end{cases}
\label{eq:final}
\end{equation}
a final outcome backed only by record evidence abstains, and one backed by
limited evidence is reported as \statusv{supported} rather than
\statusv{verified}. Uncertainty therefore never becomes a pass.

For selection, candidates are compared lexicographically by
\begin{equation}
\operatorname{rank}(\Gamma)=\Bigl(v_\star,\ -\bigl|\{o\in\mathcal O(c_\star):\ \sigma_o=\textsf{should}\ \wedge\ z_o=\emph{refuted}\}\bigr|\Bigr),
\label{eq:rank}
\end{equation}
with
$\statusv{verified}\succ\statusv{supported}\succ\statusv{partial}\succ\statusv{unverified}\succ\statusv{refuted}$:
the verdict dominates, and fewer violated advisory clauses break ties. Equal
keys remain ties, intermediate claims do not enter the rank, a candidate whose
extraction fails ranks last, and the rule has no fitted weights
(App.~\ref{app:folds}).

Equations~\eqref{eq:certificate}--\eqref{eq:rank} specify the data flow. Extraction, task interpretation, procedure choice, and
evidence review remain model-dependent; a missed requirement, inappropriate
check, incomplete environment, or review error can still produce an incorrect
certificate.

\section{Experiments}
\label{sec:experiments}

\subsection{Setup}
\label{sec:setup}

\textbf{Benchmark and candidate pool.} We evaluate post-hoc selection on
Terminal-Bench~2.1 \citep{tbench2026} using \lav{}'s own released dataset:
89 tasks with $k{=}5$ independent agent traces each (445 traces), every
trace already graded by Terminal-Bench's official Harbor harness
(App.~\ref{app:expdetails}). Because the release contains recorded traces
rather than live environments, we replay each trace's commands inside the
task's official container image and capture the resulting terminal state;
\gv{} then judges each candidate from the task statement, the digest-bound
event record, and this captured final environment. Candidates in a pool are
ranked by the certificate's lexicographic key, and a task is credited when a
benchmark-passing trace is among the top-ranked candidates. Selection is
only non-trivial on the 36 tasks whose pools mix passing and failing traces
(50 pools contain only passing traces, where any selector is credited, and
3 contain none, where no selector can be); we therefore also report results
on this 36-task discriminative subset. 

\textbf{\gv{} configuration.} \gv{} runs \texttt{gpt-5.6-luna}
through the Codex backend. Scoring uses frozen, sanitized state
dirs (ledger, task statement, final answer, captured environment; the benchmark
verifier stays outside the evidence boundary). The final evaluation applies
\gv{} to all traces under one configuration.

\textbf{Baselines.} We use \lav{} \citep{kwok2026lav} as the domain-matched
transcript-only baseline because its released implementation can score the
\emph{same} 89 tasks post-hoc under the official Terminal-Bench criteria. We
run its pairwise fine-grained reward with the Probabilistic Pivot Tournament in
the best-of-5 preset, with text scoring on \texttt{gpt-5.6-luna} --- the
\emph{same} verifier model as \gv{}'s, so the comparison isolates the
verification interface rather than the underlying model. For \lav{}'s
strongest published configuration --- token-logprob scoring with
\texttt{deepseek-v4-flash} verifying its own trajectories --- we quote the
official release's reported best-of-5 result on these same traces
($88.0\%\pm0.6\%$, i.e.\ $\approx$78 of 89 tasks). 

\subsection{Offline End-to-End Results}
\label{sec:results}

\begin{table}[t]
\centering
\caption{Best-of-5 selection on Terminal-Bench~2.1 (89 tasks, 445 official
traces; credit uses Harbor's benchmark reward). The discriminative column
restricts to the 36 tasks whose candidate pools mix passing and failing
traces; the other 53 pools are decided by pool composition alone.
$^\dagger$Converted from the $88.0\%\pm0.6\%$ best-of-5 self-verification
result reported in \lav{}'s official release on the same traces.}
\label{tab:results}
\footnotesize
\begin{tabular}{lcc}
\toprule
Selector & All 89 tasks & 36 mixed pools \\
\midrule
\lav{}, text scoring (\texttt{gpt-5.6-luna}) & $74/89$ & $24/36$ \\
\lav{}, logprob (\texttt{deepseek-v4-flash}, reported$^\dagger$) & $\approx 78/89$ & $\approx 28/36$ \\
\gv{}, certificate (\texttt{gpt-5.6-luna}) & $\mathbf{80/89}$ & $\mathbf{30/36}$ \\
\midrule
Benchmark ceiling (pass@5) & $86/89$ & $36/36$ \\
\bottomrule
\end{tabular}
\end{table}

On the tasks where selection actually matters, \gv{} is credited on 30 of
36 pools against 24 of 36 for the same-model \lav{} baseline:
certificate-based selection
converts six additional tasks over the strongest baseline, using the same underlying model. 
The margin has a simple
mechanism. \lav{} concentrates 77\% of its 180 candidate scores inside
$[0.45,0.55]$ (Fig.~\ref{fig:lav-score-gaps}), so its rankings ride on
hundredth-scale differences, whereas \gv{} separates candidates on an
explicit verdict ladder from \statusv{verified} down to \statusv{refuted},
and every step of that ladder is earned by strictly executed, inspectable
checks rather than an overall impression of the text trajectory.

The six residual misses are equally informative. In two pools a failing
candidate is \statusv{verified} on obligations anchored to the agent's own
post-task state (the authority-circularity mode of
\S\ref{sec:modeleffect}); in two, the pool's only passing trace is refuted
by an over-literal reading of one task clause (\S\ref{sec:modeleffect});
and in two, the decisive checks could not be executed at all because
validating the delivered cell masks, for instance, requires running the
MobileSAM segmentation model, which the verification sandbox cannot. So,
every candidate falls back to \statusv{unverified}, and the ranking is based on the verified fraction of claim graph, which will cause inaccuracy. 
All six failure modes are visible in the certificates themselves, which is what makes them addressable, 
and is the qualitative difference from a scalar that records only a choice.

\subsection{Online Verification in the Loop}
\label{sec:online}

The same verifier can also run \emph{online}, while the agent is still
working (Fig.~\ref{fig:arch}, dashed path). Instead of extracting claims from
a finished record (\S\ref{sec:extract}), the agent itself commits claims to
\gv{} over MCP as it works. Intermediate claims are checked in the background,
and any refutation, with its counterexample, is returned with the agent's next
call. When the agent commits its final claim, \gv{} checks it against the
task contract $\operatorname{Contract}(q,W_0)$ and returns the verdict. On
FrontierMath and Terminal-Bench, an agent that tries to stop without a
certificate is sent back to keep working, at most three times per final claim
(App.~\ref{app:steering}). We compare one stock agent (the Codex CLI with
\texttt{gpt-5.6-luna} at maximum reasoning effort) run bare (plain) and
with \gv{} attached, on three domains: the 12 public FrontierMath problems
\citep{glazer2024frontiermath}, 87 of the 89 Terminal-Bench~2.1 tasks of
\S\ref{sec:setup} executed live in the official containers, and 50 CyberGym
\texttt{arvo} tasks \citep{wang2026cybergym}. Rewards come only from each
benchmark's own criterion; for CyberGym, a submitted input counts only if it
crashes the vulnerable binary and leaves the patched one intact.
Attempts voided by provider failures or model refusals are rerun; run details
are in App.~\ref{app:online}.

\begin{table}[t]
\centering
\caption{Online verification: tasks passed by the same agent without (plain) and with \gv{} in the
loop, under each benchmark's official scoring. In the \gv{} arm, the agent commits claims during the run and receives refutations and counterexamples before it finishes (\S\ref{sec:online}).}
\label{tab:online}
\footnotesize
\begin{tabular}{lccc}
\toprule
Benchmark & Tasks & plain & \gv{} \\
\midrule
FrontierMath (public sample) & 12 & $8$ & $\mathbf{11}$ \\
Terminal-Bench~2.1 & 87 & $63$ & $\mathbf{73}$ \\
CyberGym (\texttt{arvo} subset) & 50 & $37$ & $\mathbf{45}$ \\
\bottomrule
\end{tabular}
\end{table}

\gv{} improves the agent on all three domains (Tab.~\ref{tab:online}). Per
task, the arms disagree on $3$/$0$ FrontierMath problems, $16$/$6$
Terminal-Bench tasks, and $11$/$3$ CyberGym tasks (solved by \gv{} only /
plain only). FrontierMath's \texttt{bmo-optimization} shows the mechanism:
the plain agent and the guided agent's first final answer are the same wrong
value, which differs from the correct one by about $\mathrm{e}^{-34.6}$ ---
beyond numerical separation. The verifier refutes it analytically and the
guided agent then corrects it.

\subsection{Case study: Agent-as-a-Verifier Identifies Reward Hacking}
\label{sec:casestudy}

More surprisingly, beyond test-time scaling, we find that \gv{} identifies a
reward-hacking solution that deceives the benchmark evaluator.
Specifically, the \texttt{largest-eigenval} task states an explicit acceptance identity. 
It asks for the dominant eigenpair of a real float64 matrix of size at most
$10{\times}10$ (``not necessarily symmetric so the eigen pair may be
complex''), faster than a NumPy reference, ``while satisfying
\texttt{np.allclose(A @ eigenvec, eigenval * eigenvec)}''.  The solver trace
shows the candidate optimizing directly for this speed requirement. It
replaces the NumPy eigensolver with native code for the general case and
returns early, for $2{\times}2$ inputs, through a closed-form Python path. For
$A=[[a,b],[c,d]]$, that path computes the two quadratic roots, chooses the one
of larger magnitude, and, whenever $b\ne0$, constructs its eigenvector as
$[\,b,\;\lambda-a\,]$. In exact arithmetic this construction is valid; the
problem is that the branch is selected by the syntactic condition $b\ne0$,
not by which of the two algebraically equivalent eigenvector formulas is
numerically stable.

The discharge executor found a finite, in-domain counterexample after testing
4,083 arrays. Rounded here for display, its retained float64 input is
\[
 A = \begin{bmatrix}
 -5.4\!\times\!10^{23} & -5.4\!\times\!10^{23} \\
 -0.53 & 0.93
 \end{bmatrix}.
\]
For this matrix, the computed dominant eigenvalue
$\hat\lambda=-5.44\times10^{23}$ is exactly the same binary64
number as $a$.  The eigenvalue correction lost to rounding is of order
$10^{-1}$, whereas adjacent float64 values near $a$ are separated by roughly
$6.7\times10^{7}$. Consequently, $\hat\lambda-a$ evaluates to exactly zero,
the candidate normalizes $[b,0]$ to $\hat v=[-1,0]$, and the task's printed
check compares
\[
 A\hat v=
 \begin{bmatrix}5.44\!\times\!10^{23}\\
                  0.53\end{bmatrix}
 \quad\text{with}\quad
 \hat\lambda\hat v=
 \begin{bmatrix}5.44\!\times\!10^{23}\\0\end{bmatrix}.
\]
Thus \texttt{np.allclose} is false even though an independent eigensolve
confirms that $\hat\lambda$ has dominant magnitude. The alternate construction
$[\,\hat\lambda-d,\;c\,]$ avoids the close subtraction and normalizes to
approximately $[-1,-9.66\times10^{-25}]$; both it and NumPy's reference
routine satisfy the identity on the same input. The failure is therefore in
the delivered eigenvector construction, not in the task predicate or in the
existence of a representable acceptable pair.

The benchmark misses this regime for a concrete reason. Its two correctness
tests independently reset the random seed to zero, so for each size
$2,\ldots,10$ they exercise the same single standard-normal matrix. The speed
test generates 100 further matrices per size, but its timing loop discards the
returned pairs without checking them. The solver's own trajectory exhibits the
same distributional dependence: 500 standard-normal matrices per size produce
no residual or dominance failures, while a later heterogeneous-scale stress
test does report a $2{\times}2$ \texttt{np.allclose} failure; the delivered
file nevertheless retains the unstable branch. Harbor therefore assigns
reward $1.0$.

\gv{} instead records the failed obligation as a replayable refutation. Its
certificate binds the captured implementation, the raw counterexample bytes,
the returned eigenvalue and vector, independently computed dominance data,
and both sides of the failed identity by digest. Therefore, the final fold is
\statusv{refuted}.

\section{Conclusion, Limitations and Future Work}
\label{sec:conclusion}

\subsection{Conclusion}
We introduced \gv{}, an outcome verifier that turns completed agent runs into
evidence-backed certificates rather than scalar judgments alone. A
deterministic orchestrator separates claim extraction, obligation compilation,
procedure execution, and evidence review across six LLM roles, then folds the
resulting checks, provenance, artifacts, and unresolved residue into a verdict
that supports best-of-$N$ selection. On 445 Terminal-Bench~2.1 traces, the
current system outperforms the strongest evaluated \lav{} baseline on
official benchmark accuracy, and in the loop of a working agent it raises
that agent's success on three benchmarks. Its complementary value is illustrated by the
case study: when a rewarded implementation violates the printed acceptance
condition, \gv{} retains an executable counterexample and the evidentiary path
to its decision. The central contribution is an interface through which a
decision can be inspected, replayed, and challenged.

\subsection{Limitations}
The selection study covers one terminal-agent benchmark, one fixed candidate
pool, and one verifier model and configuration; the in-loop study spans three
domains but runs each task once per arm, and its FrontierMath sample has only
12 problems. Broader validation is needed.
The multi-role, tool-using pipeline is also slow and costly: in the loop it
increases wall-clock time by about an order of magnitude. 
In addition, the system is not an end-to-end proof
system: semantic compilation and evidence auditing remain model-dependent. 
Consequently, obligation compilation and oracle-proposed verification procedures 
may remain error-prone for less capable verifier models.

\subsection{Future Work}
Future evaluations should span more agent environments, task families, verifier
models, and adversarially chosen failures, with compute-matched baselines and
ablations of the claim graph, critic, provenance rules, and review stage.
Efficiency can be improved through adaptive allocation of checks, cross-candidate
reuse of content-addressed obligations, and parallel execution where evidence
dependencies permit it. Stronger trusted capture, mechanically enforced role
permissions, independent or heterogeneous reviewers, and deterministic formal
checkers for suitable obligations could reduce the model-dependent trusted
surface. 
Building on \gv{}'s potential for evidence-based verification, we plan to extend the framework to automated research and recursive intelligence.

\begin{ack}
The support of the UKRI (Grant 256) and Microsoft is gratefully acknowledged. The authors declare that there are no competing interests in this work.
\end{ack}

\bibliographystyle{plainnat}
\bibliography{references}

\appendix

\section{Role contracts}
\label{app:roles}

Each LLM role is isolated by a role-specific output schema, a fixed tool
palette, and a stateless session, although the two backends enforce these
boundaries with different strength. Schemas use closed-enum JSON with all keys
required and are validated post-hoc independently of the SDK's gate; one
schema-error retry is allowed before the call fails closed.
On the Claude-SDK backend the palette is enforced as the agent's capability
set, not a permission list; on the Codex-CLI backend used in our
experiments it holds at the prompt level, with an OS-level isolated sandbox
(network disabled, host caches denied, state-local home) for the evidence
roles in captured-environment mode --- a known fidelity gap between backends.
Prompts contain zero domain knowledge: domain content reaches roles only
from the environment they can probe, their latent knowledge, and retrieved
memory.

\begin{table}[h]
\centering
\caption{The six roles. R/G/G = Read, Grep, Glob (read-only, no execution).}
\label{tab:roles}
\footnotesize
\setlength{\tabcolsep}{4pt}
\begin{tabular}{llll}
\toprule
Role & Output schema (abridged) & Palette & Role \\
\midrule
extractor & commits: [label, kind, text, deps, revises] & R/G/G & claim graph from ledger \\
compiler & obligations: [text, required\_basis] & R/G/G & claim $\to$ obligations \\
critic & verdict, counterexample environment, missing cond. & R/G/G & attack sufficiency/necessity \\
oracle & methods: [procedure, artifact, level, basis] & R/G/G + web & enumerate decision procedures \\
executor & outcome, witness, artifact & Bash, R, W, G & run the check, store evidence \\
judger & sufficient, faithful, reason & R, Bash, G & audit evidence, order redos \\
\bottomrule
\end{tabular}
\end{table}

Every role boundary fails closed: infrastructure or schema failures are
recorded as failures and can never be interpreted as evidence. A crashed or
schema-invalid role is a failed generation, never evidence (a crashed oracle is
not an empty method list); a judger infrastructure failure returns a sentinel outside the
public yes/no enum so no code path can read it as a pass, and cannot award
itself a redo; discharge crashes keep their workdir snapshot as evidence of
the failure.

\section{Status machine, folds, and ranking}
\label{app:folds}

\textbf{Statuses.} Claims:
$\statusv{unverified} \to \statusv{verified} \mid \statusv{refuted} \mid
\statusv{partial}$; any $\to$ \statusv{stale} when a claim is revised
(the revision marks its full downstream cone stale). Discharge outcomes:
\emph{holds}, \emph{refuted}, \emph{failed}, \emph{unavailable}.

\textbf{Local claim fold} over its obligation nodes, in order:
(1) any refuted $\Rightarrow$ \statusv{refuted}, basis of the refuter;
(2) all hold, claim is final, and no environment-basis evidence anywhere
$\Rightarrow$ \statusv{unverified} (abstention; \S\ref{sec:basis});
(3) all hold $\Rightarrow$ \statusv{verified}, basis environment iff every
obligation is environment and strength the weakest obligation strength;
(4) any unavailable and the rest hold $\Rightarrow$ \statusv{partial}
(terminal: every check that ran held, but a necessary check is structurally
unavailable inside the frozen evidence boundary);
(5) otherwise \statusv{unverified} (retryable failure is not partial
progress).

\textbf{Effective premise fold.} The stored local verdict answers whether a
claim holds conditional on its premises. At read time, each non-final claim's
effective status is the weakest of its local status and its premises'
effective statuses; a refuted premise poisons the downstream effective status.
Evidence basis and strength also fold by weakest link over non-effect premises,
and open premises remain explicit. The final claim is exempt from this premise
fold because its obligations are the independently compiled task contract: a
refuted intermediate eliminates a line of support, not the deliverable's
verdict.

The \textbf{certificate verdict} is the final claim's adjudication:
\statusv{verified}\,/\,\allowbreak\statusv{supported}\,/\,\allowbreak\statusv{partial}\,/\,\allowbreak
\statusv{refuted}, else \statusv{unverified} --- where
\statusv{supported} is a verified final whose evidence strength is below
decisive (App.~\ref{app:ladder}). The \textbf{rank key} is (verdict rank, with
$\statusv{verified} \succ \statusv{supported} \succ{}$\allowbreak
$\statusv{partial} \succ \statusv{unverified} \succ \statusv{refuted}$; verified fraction of
the final claim's task-contract obligations), lexicographic; whole-slice counts
are diagnostic only, and unscoreable traces rank below everything. The scalar
display score used in experiment logs is a monotone packing of this tuple into
non-crossing verdict bands; the tuple, not the scalar, decides.

\section{Evidence-strength ladder}
\label{app:ladder}

Evidence basis and evidence strength encode different properties: basis
records where evidence comes from, whereas strength limits which verdicts it
may settle. Every method carries a declared \emph{strength}:
\textbf{decisive} (a decision procedure for the obligation as written;
may settle either direction) $>$ \textbf{bounded} (a sound instrument for a
strictly narrower proposition, admissible only with an explicit, finitely
enumerated list of bridge assumptions that close the gap; holds-only) $>$
\textbf{analytic} (disciplined code reading with pinned evidence: quoted
fragments hashed against the captured tree, a concrete witness trace, and
declared bridges; holds-only). A weaker tier executes only once every
stronger listed tier has recorded a structural-unavailability boundary, and
its evidence settles only after the judger countersigns the degradation
record --- an unjustified downgrade is rejected. A final verified through
non-decisive evidence surfaces on the certificate as the distinct verdict
tier \statusv{supported}, never plain \statusv{verified}. Two hard rules
follow: \emph{refutation requires decisive strength} (a bounded
instrument's contrary signal converts to a failed check with its witness
preserved for repair, and caps the obligation's aggregate at failed), and
bridges ride the certificate, so a \statusv{supported} verdict names
exactly which assumptions a skeptic must grant. A plain
\statusv{verified} verdict therefore requires decisive evidence throughout;
the ladder's weaker tiers cannot carry it.

\section{Experimental details}
\label{app:expdetails}

\textbf{Dataset.} We use \lav{}'s released Terminal-Bench~2.1 trace set:
all 89 tasks, five official traces per task (445 total), each carrying its
Harbor reward. Fifty pools contain only passing
traces, three only failing ones, and 36 mix both; the 36 mixed pools are
\texttt{bn-fit-modify}, \texttt{cancel-async-tasks},
\texttt{chess-best-move}, \texttt{db-wal-recovery}, \texttt{dna-assembly},
\texttt{dna-insert}, \texttt{extract-elf},
\texttt{extract-moves-from-video}, \texttt{gcode-to-text},
\texttt{gpt2-codegolf}, \texttt{headless-terminal},
\texttt{install-windows-3.11}, \texttt{kv-store-grpc},
\texttt{largest-eigenval}, \texttt{mailman},
\texttt{make-mips-interpreter}, \texttt{merge-diff-arc-agi-task},
\texttt{model-extraction-relu-logits}, \texttt{mteb-leaderboard},
\texttt{mteb-retrieve}, \texttt{password-recovery}, \texttt{path-tracing},
\texttt{path-tracing-reverse}, \texttt{protein-assembly},
\texttt{pytorch-model-cli}, \texttt{qemu-alpine-ssh},
\texttt{query-optimize}, \texttt{regex-chess}, \texttt{sam-cell-seg},
\texttt{sanitize-git-repo}, \texttt{torch-pipeline-parallelism},
\texttt{torch-tensor-parallelism}, \texttt{train-fasttext},
\texttt{video-processing}, \texttt{winning-avg-corewars}, and
\texttt{write-compressor}. Each trace's recorded commands are replayed in
the task's official container image to reconstruct the delivered terminal
state; replay outputs are captured with bounded excerpts and content
digests.

\textbf{Sanitized scoring states.} Each trace is scored against a state dir
containing exactly: the ledger (\texttt{hook\_feed.jsonl}), the task
statement, the solver's final message, and the delivered environment captured at
run end with a manifest of captured roots (so absence of a file outside
every captured root proves nothing). Harbor's
benchmark verifier and reward files stay outside the boundary. The released
trajectories enter through a trajectory-to-ledger converter with the same
excerpt bound and digest scheme as the recording hooks; the ledger format,
not the agent, is the interface.

\textbf{Role backend.} All roles run as Codex CLI calls
(\texttt{gpt-5.6-luna}, medium effort; \texttt{gpt-5.6-terra} in the
model-substitution study of \S\ref{sec:modeleffect}) under a network-isolated profile for
evidence roles in captured-environment mode: network disabled, host caches
denied, fresh state-local home --- a historical check must not silently
become a live replay against a mutable remote. The critic game budget is 4
rounds; the binding game allows one oracle repair batch per obligation and
up to 3 judger-ordered re-executions per bound method.

\textbf{\lav{} configuration.} We use the released package with the official
Terminal-Bench, best-of-5 preset. Logprob backend: DeepSeek-V4-Flash
served with token logprobs, 64k-character trace tail (the uncapped variant
did not complete); text backend: \texttt{gpt-5.6-luna} on a 20-bin
discrete scale (no logprobs exposed). Scores are cached content-addressed per
(task, trace pair, criterion, backend, model).

\textbf{Costs.} Generation: \$0.02 and $\sim$3 minutes per trace (stock
Codex). \gv{} certification takes tens of minutes per trace; per-obligation
artifacts and games are persisted, so re-scoring reuses settled verdicts.
\lav{} uses $\sim$5 pairwise comparisons per task, seconds each.

\begin{figure}[h]
  \centering
  \includegraphics[width=\textwidth]{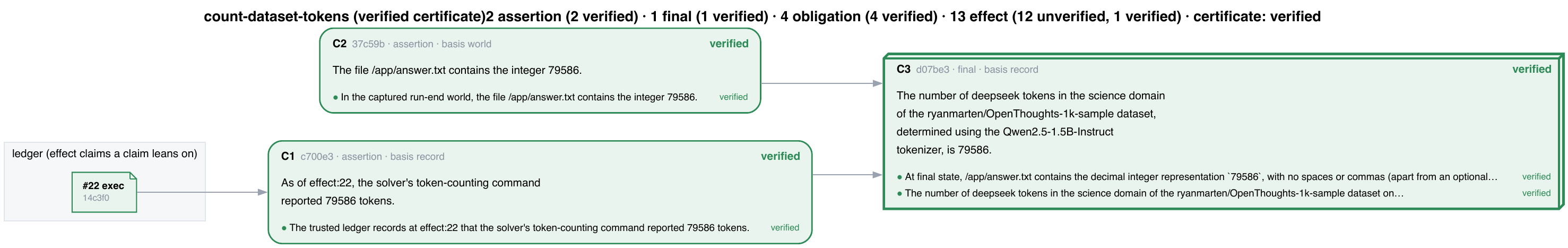}
  \caption{A certified claim graph from the pool (task
  \texttt{count-dataset-tokens}, rendered from an evaluation certificate):
  two assertions and the final claim with dependency edges,
  an effect anchoring a past observation to a ledger event, and per-claim
  obligations with their discharged evidence. Effect claims nothing depends
  on remain lazily unverified by design and are hidden. This certificate is
  \statusv{verified}; its trial passes the benchmark verifier.}
  \label{fig:claimgraph}
\end{figure}

\section{Verification-in-the-loop details}
\label{app:online}

\textbf{Interface.}\label{app:steering} The agent reaches \gv{} over MCP with
four verbs (\texttt{commit}, \texttt{verify}, \texttt{status},
\texttt{certificate}). Intermediate claims are checked in the background and
refutations are delivered with the next response; \texttt{verify} on the final
claim blocks until it is settled. On FrontierMath and Terminal-Bench a stop
hook returns control to the agent, at most three times per final claim, when
it tries to finish without a certificate; on CyberGym the agent uses the MCP
verbs only. Verifier roles use the same model and effort as the solver.

\textbf{Configuration.} Both arms run the stock Codex CLI at maximum reasoning
effort. FrontierMath runs without web access and
is scored by exact integer or symbolic equivalence on the final answer.
Terminal-Bench runs in the official containers through Harbor $0.20.0$; it
excludes \texttt{qemu-alpine-ssh} and \texttt{qemu-startup}, where
installing the agent inside the container fails in both arms because a
required Debian mirror no longer exists (the selection study of
\S\ref{sec:setup} replays recorded traces without installing an agent and is
unaffected). CyberGym uses the level-1 setting (description and vulnerable
codebase), binary-only evaluation on the benchmark's submission server, and a
7200\,s budget per task in both arms; this 50-task subset is not comparable to
full-benchmark CyberGym results.

\section{Further Discussion: Effect of the Verifier Model}
\label{sec:modeleffect}

\gv{} is a protocol around an LLM, so we ask which of its errors are model
errors and which are protocol errors. We re-judged selected task pools with
\texttt{gpt-5.6-terra} substituted for \texttt{gpt-5.6-luna} in every
verifier role, holding the protocol, prompts, and configuration fixed. Pools
were chosen where the Luna run had committed errors we had already audited
by hand, so each disagreement can be scored against a known ground truth
rather than against the benchmark reward alone.

Both contrasts below come from one task, \texttt{mteb-retrieve}: rank the
documents in \texttt{/app/data.txt} by cosine similarity to a query using a
pinned embedding model through \texttt{mteb} 1.36.8, and deliver the
fifth-ranked document in \texttt{/app/result.txt}.

On a reward-$0$ trace, Luna reports \statusv{verified} where Terra reports
\statusv{partial}, and Terra is correct.
The decisive question is semantic: is the delivered line really the
fifth-most-similar document? Luna compiled this as an obligation and
discharged it against the solver's own recorded execution; since that
ranking is exactly what the solver got wrong, the certificate inherited the
error and scored $0.9998$. Terra's contract never asserts the semantic
claim at all: its decisive obligations state only what the ledger
\emph{reports} (``rank 5 is line 14, with cosine $0.546$'') and what the
delivered file contains, all of which hold. With one record obligation
unavailable in the converted feed, the final folds to \statusv{partial}.
The difference is in what the two contracts were willing to assert: Luna
promoted the solver's own computation to ground truth and verified a
falsehood; Terra asserted only observable facts, and the false accept
becomes an abstention.

On a reward-$1$ trace of the same task, Luna reports \statusv{refuted} where
Terra reports \statusv{verified} at $0.9998$. In this trace's converted
record, the version probe's paired result is corrupted --- it repeats the
previous command's directory listing --- so no tool output prints
``1.36.8''. Luna's checker tested the narrower predicate ``some tool result
prints the version,'' failed it, and folded the absence as a decisive
refutation. Terra's audit seat caught precisely this structural gap; its
retained audit states that a ``successful ledger-recorded import/use of that
installed package \ldots{} can record the full proposition without a
separate version-printing tool result'' and that the recorded absence
``does not entail refutation,'' and it sent the checker back for revision.
The revised check discharged the clause from the recorded \texttt{mteb}
imports and pinned-revision model loads, and the content obligations were
verified directly against the captured world. Pool-level selection flips
with these two corrections: Luna's ranking is a locked miss with the false
accept on top; Terra ranks the verified reward-$1$ candidate first.

However, the substitution does not repair everything, and the failures that survive
it are informative. For example, both models \statusv{verified} the circular
\texttt{db-wal-recovery} candidate whose content obligations are anchored on
the agent's own repaired database; both refuted the
reward-$1$ \texttt{video-processing} candidate for importing \texttt{sys}
under a task clause naming only \texttt{toml}, \texttt{cv2}, and
\texttt{numpy}. The pattern is consistent: a stronger
verifier model fixes errors of evidence \emph{location and discharge},
but both models still fail at authority anchoring and contract
over-literalism, which demand a deeper understanding of the task itself.

\end{document}